\documentclass{article}

\usepackage[english]{babel}

\usepackage[letterpaper,top=2cm,bottom=2cm,left=3cm,right=3cm,marginparwidth=1.75cm]{geometry}

\usepackage{amsmath}
\usepackage{graphicx}
\usepackage[colorlinks=true, allcolors=blue]{hyperref}
\usepackage{authblk}
\usepackage[authoryear]{natbib}

\title{LLM and Human Modes of Representation}
\author{Shalom Lappin}
\affil{Queen Mary University of London and\\
        King's College London\\
        \texttt{s.lappin@qmul.ac.uk}}
        
\begin{document}
\maketitle

\begin{abstract}
Much work on the cognitive foundations of AI has focussed on comparisons between the ways in which Large Language Models (LLMs) 
and humans process information and represent it. One aspect of this comparison involves determining the extent to which LLMs can achieve 
or surpass human performance on a variety of cognitively interesting tasks. A second explores points of convergence and divergence between 
LLM and human systems for processing information. Here, I consider some recent research that has addressed both issues in two informational
domains. The first is the representation of linguistic knowledge. The second is real world reasoning and planning. While LLMs frequently achieve 
impressive levels of performance and fluency on linguistic applications, they tend to handle linguistic content in ways that are distinct from human 
processing. They are also, for the most part, less efficient than humans in learning and generalisation for reasoning tasks. 
\end{abstract}

\section{Introduction}

LLMs have achieved a remarkable degree of competence across a variety of cognitively interesting tasks, often surpassing human performance. This has created 
a polarising debate. One strand of opinion has hailed the models as the breakthrough phase on the way to General Artificial Intelligence. An  opposing view dismisses 
them as no more that "stochastic parrots" that reproduce the data on which they are trained.\footnote{See \cite{Lappin2024,Lappin2025a} for an attempt to assess the 
  actual capacities of LLMs, with critical comment on a range of extreme views of this kind.} In order to achieve a factually grounded perspective on what LLMs can do, 
it is necessary to study how they operate, comparing them to human performance on the same applications.

In this article I look at two central domains of human learning and representation. Section \ref{ling_rep} considers linguistic knowledge, while Section \ref{reasoning} 
addresses reasoning and planning. In these Sections I compare human with LLM performance for five tasks, and I consider LLM performance on a sixth task in
the second Section. Subsection \ref{sent_accept} focusses on the way in which humans and LLMs rate sentence acceptability. It presents an overview of research 
on how humans and LLMs perform on this task, both within and independently  of different kinds of context. Subsection \ref{syntactic_hierarchy} explores the extent to 
which LLMs recognise hierarchical  syntactic structure as a condition for identifying subject-verb agreement. It summarises some recent experimental work on LLM  
and human performance for the agreement task in sentence structures that exhibit increasing complexity. Subsection \ref{narrative} cites some new experimental 
work that compares human and LLM generated texts describing visual sequences. 

LLMs do well on these tasks, converging on human levels of performance (in the case of agreement, exceeding it). However, They process and represent 
linguistic information in ways that are strikingly different than those exhibited by humans.

The first reasoning task which I take up is natural language inference (NLI), in Subsection \ref{nli}. It has been extensively studied, and reported on in the NLP 
literature.  LLMs show high levels of performance compared to humans, for in domain test sets, but these quickly decline with out of domain testing. This suggests
that LLMs are applying powerful pattern recognition techniques rather than deeper reasoning to perform NLI tasks. A similar conclusion emerges from recent work 
on complex image interpretation (Subsection \ref{image}), and planning (Subsection \ref{planning}). 

\cite{Mahowald&etal2024} provide evidence that LLMs achieve a high level of performance for purely linguistic applications. They describe this sort of knowledge
as \emph{formal linguistic competence}. However, they show that LLMs are far less capable of dealing with real world inference and reasoning tasks in natural 
language. Mahowald et al. term this latter ability \emph{functional linguistic competence}.  They argue that in humans these two competences are encoded in 
distinct neural mechanisms. The experimental work that I consider here supports this broad distinction. But it  indicates that humans and LLMs also use different 
processing and representational methods for \emph{formal} linguistic tasks.

\section{Representing Linguistic Knowledge}
\label{ling_rep}

\subsection{Sentence Acceptability}
\label{sent_accept}

Distinguishing well-formed from ill-formed sentences is a basic component of a native speaker's knowledge of his/her language. Theoretical linguists often
associate this ability with the identification of the set of grammatical sentences, and its complement, the set of ungrammatical sentences. \cite{Lau&etal2017} 
point out that grammatically is a theoretical property, which cannot be directly observed. Speakers have access to the observational property of naturalness, which 
corresponds to acceptability. They demonstrate, through a series of experiments, that sentence acceptability is a gradient property, in contrast to binary classifiers, 
like odd vs even numbers. Competent native speakers are able to reliably apply both types of classifier.\footnote{I leave open the question of whether or not 
  grammaticality is a binary property, and how to characterise it. In principle, it could be specified as gradient, analagously to acceptability. In that case we 
  could use probability based  metrics to predict degree of grammaticality. See \cite{Lau&etal2017} for discussion, and for references to alternative views on 
  this question.}

\cite{Lau&etal2017} use crowd sourcing to collect speaker acceptability ratings for sentences in several languages. Their test set for each language includes well-formed 
sentences, and sentences with infelicities introduced by round-trip machine translation through a variety of target languages. They experiment with several types of machine
learning models to predict aggregate human ratings for these sentences. They find that a recurrent neural network (RNN) did best in predicting human acceptability judgments,
as measured by Pearson correlation, achieving a Pearson value of 0.53 . However, it was necessary to map the logprob values that the models assigned to the sentences into 
acceptability scores, with scoring functions, in order to obtain reasonable correlations. The scoring functions effectively filtered out sentence length and lexical frequency as 
noise factors. The most robust and successful of these functions was SLOR

\begin{center}
{\Large{
$SLOR~~~~~~~~\dfrac{log~P_m(S)  - log~P_u(S)}{|S|}$ 
}}
\end{center}

\noindent
where $S$ is a sentence, $P_m(S)$ is the probability that the model $m$ assigns to $S$, and $P_u(S)$ is $S$'s unigram probability in $m$.\footnote{\cite{Pauls&Klein2012} 
  first introduced SLOR for a another task.}
  
While \cite{Lau&etal2017} study human sentence acceptability ratings independently of context, \cite{Lau&etal2020} consider them in various types of document context.
They also experiment with first generation transformers as models for predicting these ratings. The best performing models are BERT and XLNET, both bidirectional, which 
achieve Pearson correlations of 0.70-0.74 in their predictions of human ratings. The optimal scoring function for these models is not SLOR, but PenLP

\begin{center}
{\Large{
$PenLP~~~~~~~~~~~\dfrac{log~P_(S)}{((5 + |S|)/(5 + 1))^\alpha}$ 
}}
\end{center}

\noindent
    where $\alpha = 0.8$.  PenLP has been widely used for beam search decoding in machine translation \cite{Vaswani&etal2017}.  As a scoring function for mapping logprob
values into acceptability scores it provides a heuristic device for filtering sentence length.  

\begin{figure}
\centering
\includegraphics[scale=0.9]{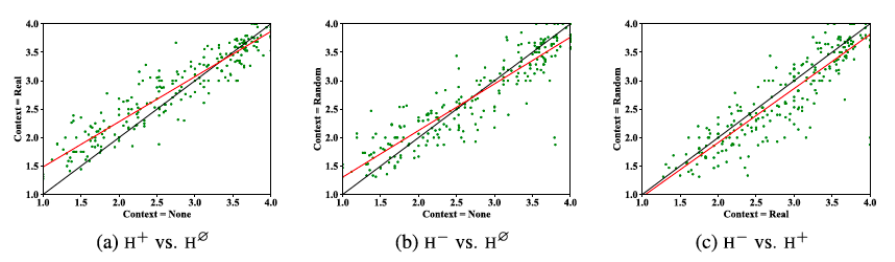}
\caption{Human Sentence Ratings in Textual Contexts, Lau et al. (2020). H$^+$ = human ratings in a relevant context. H$^-$ = human ratings in an irrelevant context.
              H$^\emptyset$ = human ratings in a null context.}
\label{lau-etal20}
\end{figure}

\cite{Bernardy&etal2018} and \cite{Lau&etal2020} observe an interesting \emph{compression effect} on human sentence acceptability judgments in textual contexts. This effect involves 
raising ratings at the low end of a four point scale, and lowering them at the high end. It is present both in textual contexts that are relevant to the sentence, and those which are
irrelevant. Figure \ref{lau-etal20} from  \cite{Lau&etal2020} illustrates this pattern. The red line is the regression fit for  the human ratings. The graphs in Figure \ref{lau-etal20} 
(a) and (b) exhibit the compression effect.\footnote{\cite{Lappin2021} presents total least square regression analyses, performed by Jey Han Lau, which demonstrate that
  the compression effect in this data is robust, and it is not the result of regression to the mean.} 
  
 They explain this effect as the result of two factors: cognitive load and discourse coherence. The first takes up processing resources to interpret a preceding text, and so it
 operates both in contexts that are relevant to a following sentence, and in those that are not. Cognitive load causes judgements to move toward the centre of the rating scale
 because of less attention being available to focus on the form and content of the sentence. Discourse coherence improves the ratings of sentences with infelicities when
 the preceding textual context has a clear connection to it, and helps to clarify its content. Therefore it only applies in relevant contexts. This would account for the fact that
 raising at the lower end of the rating scale is larger, and lowering smaller at the higher end, in Figure \ref{lau-etal20} (a) than in Figure \ref{lau-etal20} (b).
 
\cite{Jang&etal2026} examine the influence of visual contexts on human and vision LLM (VLM) sentence acceptability ratings. They use a similar experimental protocol to 
the one that  \cite{Lau&etal2020} employ. They construct a test set of well-formed sentences, and sentences containing infelicities introduced through machine translation. 
The sentences are taken from online materials produced after the training dates of the LLMs that they examine, to avoid contamination of the test set. They use GPT-5 to 
generate images that are relevant to sentences, and they  randomly select from them to obtain irrelevant ones for each sentence. 

\cite{Jang&etal2026} solicit human acceptability judgments through Prolific, for the sentences in the test set, under each of the visual context conditions, relevant, irrelevant, 
and null. They also test a closed source LLM, ChatGPT-4o (mini and large), and three open source models, InternVL3 (1B and 8B parameter variants), Qwen2.5 (3B and 7B 
parameter variants), and llava-1.5 (7B parameters). 

With the exception of llava-1.5, the LLMs predict human sentences rating, under prompt, with a high Spearman correlation. Interestingly, the best scores were
for the null visual context ratings, with ChatGPT-4o receiving 0.89, InternVL3-8B 0.88, and Qwen2.5-7B 0.84. llava-1.5 did poorly (0.39), but this appears to be due to the fact
that it was not trained as a prompt following model. The normalised logprob values that each of the open source models, including llava-1.5, assigns to the sentences,
show high Spearman correlations to human ratings. They run from 0.72 (InternVL3-8B) to 0.79 (Qwen2.5-7B), with the null context generally giving the best results. 

\begin{figure}
\centering
\includegraphics[scale=1.1]{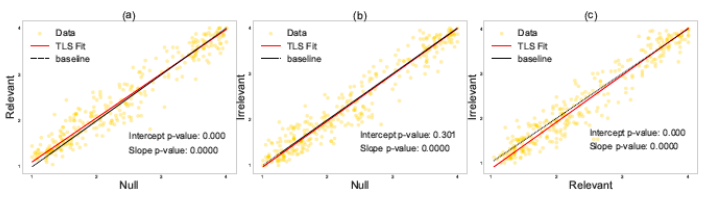}
\caption{Human Sentence Ratings in Visual Contexts, Jang et al. (2026)}
\label{jang_hum26}
\end{figure}

\begin{figure}
\centering
\includegraphics[scale=1.1]{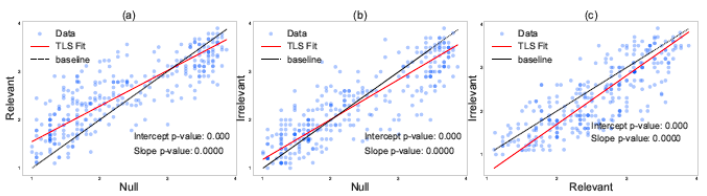}
\caption{Qwen2.5-7B Sentence Ratings in Visual Contexts, Jang et al. (2026)}
\label{jang-qwen26}
\end{figure}

One of the most striking outcomes of \cite{Jang&etal2026}'s experiments is their finding that the compression effect is (more or less) absent from human acceptability 
judgments in visual contexts, but it does appear in LLM ratings for these cases. We can see this contrast in the difference between \cite{Jang&etal2026}'s graph for 
the human ratings given in Figure \ref{jang_hum26}, and their graph for Qwen2.5's ratings in Figure \ref{jang-qwen26}. The latter resembles \cite{Lau&etal2020}'s 
graph for human sentence ratings in textual contexts, shown in Figure \ref{lau-etal20}.

\cite{Jang&etal2026} suggest that textual context provides introductory content for a sentence which humans rate, and hence it is difficult to disregard in the course
of processing the sentence. However, images are accessed through a different modality, and so they are easier to filter out when interpreting the sentence. LLMs retain 
the image as part of the extended context of the sentence, and so it influences the model's processing of the sentence in a way in which it does not for humans. The 
image is factored into the output vector that the model produces for the sentence that follows it, because VLMs are trained to process contexts in this way. Humans 
appear to be more selective in what they retain from different modalities, and they tend to suppress the preceding image when assessing a sentence for acceptability.

\subsection{Hierarchical Syntactic Structure}
\label{syntactic_hierarchy}

The sentences in 1(a)-(d), from \cite{Lappin2021}, illustrate the fact that subject-verb agreement depends on hierarchical syntactic structure, rather than on the linear order
of the subject NP relative to the verb with which it agrees.

\begin{enumerate}
  \item[1(a)] \emph{The students submit} a final project to complete the course.
  \item[(b)] \emph{The students} enrolled in {\bf{the program}} \emph{submit} a final project to complete the course.
  \item[(c)] \emph{The students} enrolled in {\bf{the program}} in {\bf{the Department}} \emph{submit} a final project to complete the course.
  \item [(d)] \emph{The students} enrolled in {\bf{the program}} in {\bf{the Department}} where {\bf{my colleague}} teaches \emph{submit} a
                  final project to complete the course.
\end{enumerate}

\noindent
Each of the NPs in bold in 1(b)-(d) is closer to the verb \emph{submit}, but more deeply embedded than the head of the subject NP, \emph{The students},
which controls this verb.

\cite{Linzen&etal2016} show that a a Long Short Term Memory (LSTM) RNN can recognise subject-verb agreement relations to a high degree of accuracy,
but success declines as the number of intervening \emph{distractor} NPs increases. \cite{Bernardy&Lappin2017} experiment with different sorts of RNNs and 
Convolutional Neural Networks (CNNs), and they extend the number of embedded distractor cases. They confirm \cite{Linzen&etal2016}'s basic finding that
LSTMs, as well as Gated Recurrent Units (GRUs), achieve reasonable results, with a decline in performance in correlation to the number of distractors.\footnote{See
  \cite{Lappin2021} for detailed discussion of this, and other work on agreement, within an LSTM framework.}  This pattern approximates human performance, 
  for which difficulty in processing these structures increases with the number of distractor NPs, and the depth of their embedding. 
  
In more recent work \cite{Lakretz&etal2022} test transformers on agreement within short and long recursive nested constructions. 2(a) is an example of a
short recursive nested construction, where the embedded subject NP is adjacent to the verb with which it agrees. 2(b) is an instance of a long nested
structure in which the embedded subject is separated from its verb by a PP modifier that it contains.

\begin{enumerate}
  \item[2(a)] \emph{The keys} that {\bf{\emph{the man holds}}} \emph{are} for his car.
  \item[(b)] \emph{The keys} that {\bf{\emph{the man}}} near the cabinet {\bf{\emph{holds}}} \emph{are} for his car. 
\end{enumerate}

\cite{Lakretz&etal2022} experimented with five variants of GPT-2, and two versions of RoBERTa. They found that while  all the transformers achieved near 
perfect accuracy for the short nested sentences, they dropped to below chance for the long structures. This sharp decline was produced by the addition of 
a three word PP modifier between the embedded subject and its verb. Humans generally perform better on long nested structures like 2(b).

\cite{Lampinen2024} observes that while human subjects for agreement tasks are given instructions, \cite{Lakretz&etal2022}, and others who have reported
similar results for transformers, did not prompt the models that they tested. He experiments with two versions of the LLM Chinchilla, one with 7B parameters, 
and one with 70B. He uses two example sentence (two-shot) prompts, and eight example sentence (eight-shot) prompts, to test them on a variety of short and 
long nested agreement dependencies, with different levels of embedding complexity. 

He finds that even with the two-shot prompt the small model performs at least as well as humans, and in more complex nested cases it surpasses them. The 
large model outperforms humans in almost all of the nested agreement structures. With eight-shot prompting both models improve significantly.  He notes that 
the two-shot prompt contains fewer examples and less information than the instructions given to human subjects in earlier experiments on nested agreement. 

\cite{Oh&Linzen2025} argue that current LLMs generally surpass human performance on most linguistic tasks. This is due to their capacity to produce large 
context vectors in which they have perfect access to large amounts of information. These context vectors facilitate the recognition of dependencies among 
disparate elements of linguistic input, over large, complex structures. Memory limitations render human processing of such dependencies increasingly difficult,
in relation to the size and the structural complexity of the sequence of tokens that intervenes between the elements of the dependence. \cite{Lampinen2024}'s 
experiments support this view. Similarly \cite{Jang&etal2026}'s results suggest that the contrast between human and LLM  responses to visual image contexts 
in the sentence acceptability rating task, reflects the fact that the models retain visual information that humans filter out or suppress. 

Some of the differences between the ways in which humans and transformers process linguistic information may derive from the fact that LLMs have more powerful 
resources for retaining content over time. Humans rely on sophisticated filtering and suppression techniques to render processing tasks manageable with more 
limited memory and computing mechanisms.

\subsection{Generating Narrative}
\label{narrative}

LLMs are particularly successful in generating fluent, natural sounding text in response to queries and prompts. It is often difficult to distinguish this text
from human writing, without additional information on the author(s). This development has produced a widespread tendency to anthropomorphise
LLMs, and to attribute animate agency to them, sometimes with unfortunate consequences. It has also flooded the internet with machine generated content, 
much of it of containing false information and fabricated claims. Email scams, dubious scientific articles, and fraudulent news reports have flourished through
the ability of LLMs to mass produce credible text, as well as realistic images, and audio reproduction.  

\cite{Ilinykh&etal2026} compare human stories for film image sequences, with the visually grounded narratives that a set of current LLM models generate 
for the same sequences.  They specify five quantitative metrics for measuring different dimensions of coherence in discourse. These include coreference 
continuity, typology of implicit discourse relations, topic change, character persistence, and visual grounding of character references. Each of these metrics 
corresponds to a distinct aspect of narrative coherence. So, for example, if a discourse features a wider range of implicit discourse relations, and greater 
persistence of characters, it will generally display a higher level of coherence. 

They apply these metrics to 60 human stories in \cite{Hong&etal2023}'s data base of crowd sourced stories for film clips, and to the narratives that five LLMs 
generate for the same clips.  The LLMs are CLAUDE 4.5 and ChatGPT-4.o, for the closed source models, and INTERNVL3, LLAMA 4 SCOUT, and QWEN3-VL, 
for the open source models. The models received the same instructions, formulated as a prompt, that \cite{Hong&etal2023} used for their human subjects in the 
crowd source task. Humans score higher than the LLMs on all but the visual character grounding metric. This seems to reflect the fact that human stories
are less directly tied to the details of the images in the sequence for which they construct a story. 

\cite{Ilinykh&etal2026} specify a single integrated coherence metric by aggregating the five individual values of each of the measures that they test. The unified
metric is given in two variants, as the arithmetic mean and the geometric mean. The latter tends to give a lower score to cases in which there is a significant
imbalance among the constituent values from which it is computed. The human narratives received the highest integrated coherence score on both versions
of the metric. 

\cite{Ilinykh&etal2026} experiment with  a longer prompt, which forces attention to each image in a sequence, and encourages more detailed and elaborate
narratives for the sequence. They used this prompt to crowd source a new set of human stories through Amazon Mechanical Turk, and they applied it to the
five models. The metrics indicate that the models tended to converge on human performance for some aspects of coherence, but not in others. The human 
narratives still received the highest unified score, in both variants, with the more elaborate prompt. These scores are given in Tables \ref{ncs_arithm} and
\ref{ncs_geom}.

\begin{table}[t]
\centering
\begin{tabular}{l|c|c}
  System & Short Prompt & Long Prompt\\
  \hline
  \texttt{CLAUDE} & $0.42$ & $0.41$\\
  \hline
  \texttt{GPT-4o} & $0.45$ & $0.46$\\
  \hline
  \texttt{INTERNVL3}  & $0.43$ & $0.40$\\
  \hline
  \texttt{LLAMA 4 SCOUT} & $0.32$ & 0.$38$\\
  \hline
  \texttt{QWEN3-VL} & $0.45$ & $0.41$\\
  \hline
  \texttt{HUMANS} & ${\bf {0.50}}$ & ${\bf{0.48}}$\\
  \hline
\end{tabular}
\caption{Narrative Coherence Score (NCS) Arithmetic Mean, Ilinykh et al. (2026)}
\label{ncs_arithm}
\end{table}

\begin{table}[t]
\centering
\begin{tabular}{l|c|c}
  System & Short Prompt & Long Prompt\\
  \hline
  \texttt{CLAUDE}  & $0.29$ & $0.26$\\
  \hline
  \texttt{GPT-4o} & $0.32$ & $0.33$\\
  \hline
  \texttt{INTERNVL3} & $0.30$ & $0.26$\\
  \hline
  \texttt{LLAMA 4 SCOUT} & $0.06$ & $0.17$\\
  \hline
  \texttt{QWEN3-VL} & $0.32$ & $0.27$\\
  \hline
  \texttt{HUMANS} & ${\bf {0.37}}$ & ${\bf{0.37}}$\\
  \hline
\end{tabular}
\caption{Narrative Coherence Score (NCS) Geometric Mean, Ilinykh et al. (2026)}
\label{ncs_geom}
\end{table}

In an additional experiment \cite{Ilinykh&etal2026} compare the human and the model generated stories for perplexity. They apply each of the open source
models to measure the perplexity of the narratives produced by the humans, and by the other models, for both prompts. They also analyse randomly scraped 
online human text, for comparison. The human text displayed substantially higher perplexity than the models' narratives, under both prompts, across all 
models' measurements. The perplexity values are shown in Table \ref{perplexity}.

\begin{table}[t]
{\small{
\centering
\begin{tabular}{l|c|c|c|c|c}
  Evaluator & Human$_{SD}$ & Human$_{SP}$ & Models$_{SP}$ & Human$_{LP}$ & Models$_{LP}$\\
  \hline  
  \texttt{QWEN3-VL} & ${\bf{14.21}}$ & ${\bf{13.58}}$ & $3.12-4.31$ & ${\bf{11.54}}$ & $2.67-4.11$\\
  \hline   
  \texttt{LLAMA 4 SCOUT} & ${\bf{37.00}}$ & ${\bf{25.98}}$ & $1.83-18.04$ & ${\bf{20.85}}$ & $1.67-14.42$\\
  \hline
  \texttt{INTERNVL3} & ${\bf{9.39}}$ & ${\bf{11.44}}$ & $2.41-5.93$ & ${\bf{10.30}}$ & $3.39-5.05$\\
  \hline
\end{tabular}
\caption{Human text and model text perplexities. $SD$ = scraped data, $SP$ = short prompt, and $LP$ = long prompt, Ilinykh et al. (2026)}
\label{perplexity}
}}
\end{table}

The combined results of this work suggest that human narrative is both less predictable (it has a higher surprisal level) and more cohesive than LLM generated 
text. This may provide experimental grounding to the intuition that human discourse is more original, and reflects a deeper pattern of understanding, than the 
narratives that models generate. LLM language is fluent on the surface, but it seems to lack some of the distinctive features of human speech and writing. 

\newpage
\section{Reasoning}
\label{reasoning}

\subsection{Natural Language Inference}
\label{nli}

Natural Language Inference (NLI) is a reasoning task that involves classifying the relation between a premiss and a conclusion as entailment,
contradiction, or neutral. These relations depend on the lexical semantic content of the premiss and the conclusion, as well as real world knowledge.
The examples in 3 are taken from \cite{Bowman&etal2015}'s Stanford NLI (SNLI) annotated corpus.

\begin{enumerate}
  \item[3(a)]  A soccer game with multiple males playing. $\rightarrow$\\
    Some men are playing a sport.\\
    {\bf{Entailment}}
  \item[(b)] A man inspects the uniform of a figure in some East Asian country. $\rightarrow$\\
    The man is sleeping\\
      {\bf{Contradiction}}  
  \item[(c)] An older and younger man smiling. $\rightarrow$\\
    Two men are smiling and laughing at the cats playing on the floor.\\
       {\bf{Neutral}} 
\end{enumerate}

LLMs that are fine-tuned on corpora of this kind generally achieve human levels of performance when tested on hold out cases from the corpus, or on data 
the same kind. However, \cite{Talman&Chatzikyriakidis2019} show that BERT's accuracy declines significantly in out of domain testing, where the test
inferences are distinct from those in the training set. 

\cite{Talman&etal2021} do adversarial testing on BERT. They fine-tune it on corrupted training sets containing meaningless sentences and test it on
meaningful inference pairs. They also go in the converse direction by fine-tuning BERT on well-formed inferences, and testing it on corrupted 
pairs. In both cases BERT achieved unreasonably high scores, given the distortions in either its training or its test data. \cite{Talman&Chatzikyriakidis2019}'s
results indicate that BERT has difficulty in generalising inference recognition to new cases that are substantially different than those that it encountered
in training. \cite{Talman&etal2021}'s experiments suggest that BERT is relying primarily on surface pattern matching and statistical cues to classify NLI 
relations, rather than deeper lexical semantic reasoning. 

In more recent work \cite{Sadat-Caragea2024} construct a labelled scientific NLI training and test set, from sentence pairs in five computer science 
journals, each representing a distinct subdomain of the field. They test four transformers fine-tuned on the training part of the corpus, BERT, SciBERT, 
XLNET, and RoBERTa. They also test two LLMs, Llama-2 and Mistral, with zero shot, and few shot prompt learning. 

Their best performing model is RoBERTa, with an overall F1 score of 77.2 for all subdomains of the test set. The other three fine-tuned transformers
score between 73.24 and 76.48. The two LLMs have far lower levels of performance, with Llama-2 obtaining the highest overall F1 of the two models,
at 51.77 on three-shot prompt learning. In out of domain testing through subdomain shift, RoBERTa's F1 declined slightly by $\sim$ 2\%.

\cite{Sadat-Caragea2024} test expert and non-expert human subjects by having them re-annotate randomly selected subsets of the test set. They then 
estimate their performance on the entire set. The experts outperform RoBERTa by a significant margin, with an F1 score of 89.33. The non-experts still 
score higher, than RoBERTa, with a reduced advantage, at 79.78.

While fine-tuned transformers and prompt guided LLMs can achieve reasonable results for in domain tested NLI tasks, they are still at a significant distance 
from human performance on domain general inference that depends on lexical semantic and real world knowledge, particularly across diverse areas of 
specialised knowledge.  

\subsection{Complex Image Interpretation}
\label{image}

Deep neural networks have been successfully applied to images whose interpretation requires understanding of the properties and relations among objects 
that appear in the images. \cite{Cunnington&etal2023}, for example, propose a hybrid model for identifying images whose recognition requires understanding 
the rules of a game. It consists of a CNN, enriched by an Inductive Logic Programming component that learns rules concerning features that the CNN extracts
from photographs.\footnote{See \cite{Lappin2025b} for a discussion of neuro-symbolic models of this kind.}

\cite{Albaqshi&etal2025} test several LLMs on the diagnostic interpretation of neurological images (CT scans and MRIs). They experiment with GPT, Gemini 1.5 
pro, Gemini Flash, Claude 3.0, and Claude 3.5, on a set of diagnostic quizzes  in neurological journals, containing images and textual descriptions. They control 
for contamination of the test set by rephrasing the quiz texts. They also experiment with text only versions of the quizzes, in which the images are omitted,
They compare their models to three human medical experts, of different levels of experience. 

\cite{Albaqshi&etal2025} find that Claude 3.5 is the best performing model across all variants of the quizzes, obtaining 76.8\% on both the rephrased text + image
quiz, and the rephrased text only quiz. It outperforms two of the three human experts. The other models score from the low 50s through to 75\% on the two quiz 
variants. Interestingly, with the exception of Gemini Flash, the difference between the models' accuracy on the rephrased text + image quiz and the rephrased 
text only quiz is not statistically significant. As \cite{Albaqshi&etal2025} observe, this suggest that for the most part, the models are relying on descriptions of the
symptoms for their diagnostic conclusions. They also report that the models have limited success in identifying the location of pathologies in the images,

\cite{Zhou&etal2026} Experiment with OpenAI 04-mini-high, Claude 4 Opus, Gemini 2.5 Pro, and Qwen 3 on 200 neurological image quizzes from the 
\emph{New England Journal of Medicine}. Unlike \cite{Albaqshi&etal2025}, they do not control for contamination of the test data, and they acknowledge
that their models may have seen some of it in their training. Therefore, the results that they report must be treated with caution. 

They report that OpenAI 04-mini-high outperformed the other models, and with scores of over 90\% in text + image, text only, and image only versions of the quizzes, 
and across all levels of difficulty. It also scored higher than four human experts, each having a different degree of experience. They confirm \cite{Albaqshi&etal2025}'s 
finding that, in general, the presence or absence of the image does not affect the accuracy of the models, which can achieve the same score on the basis of textual 
descriptions of symptoms. They also note that error analysis of the results for OpenAI 04-mini-high indicate that it's mistakes were not due to problems of image 
identification, but to faulty diagnostic reasoning.

The general conclusion that seems to follow from this work is that while LLMs are powerful instruments for diagnostic reading of medical images, their success is
driven by an enhanced capacity for pattern recognition. They are not skilled at deep diagnostic reasoning, unless they are specifically fine-tuned to do it, through
exposure to domain specific information, and extensive feedback in the form of prompting. 

\subsection{Planning}
\label{planning}

Planing involves determining a sequence of actions for achieving clearly specified goals. Devising optimal procedures for efficiently implementing these strategies 
poses substantial challenges of reasoning and inference. This has been one of the core areas of AI research since its emergence as a field of study. Deep Neural 
Networks (DNNs) have been highly successful in solving planning problems in specific domains, such as mastering winning strategies for complex games. Google DeepMind's
\emph{AlphGo} (\cite{Silver&etal2017}) defeated one of the top grandmasters of Go in 2017.  \emph{AlphGo} is a CNN trained on Go moves and game plans, with 
Reinforcement Learning. 

LLMs have achieved success on domain specific planning tasks, but it is unclear to what extent they are capable of the general reasoning required for zero, or few
shot transfer to out of domain strategy problems. \cite{Duchnowski&etal2025} examine the performance of several LLMs on three NP-hard optimisation problems
involved in planning tasks. Two of these are graph navigation tasks, and one is a variant of a bin packing problem. To solve the graph colouring task one must
assign distinct colours to all the adjacent nodes of an undirected graph, with the minimum number of colours. In the travelling salesman problem one searches 
for the shortest route through all the nodes of a graph, without visiting any node more than once. For the knapsack task one must fill the sack with a maximal 
number of objects, whose total weight is below a given threshold. These problems are NP-hard because the number of solutions that must be considered is 
exponential relative to the number of items in the task (nodes in the graph, and objects to be placed in the knapsack). 

\begin{figure}
\centering
\includegraphics[scale=0.9]{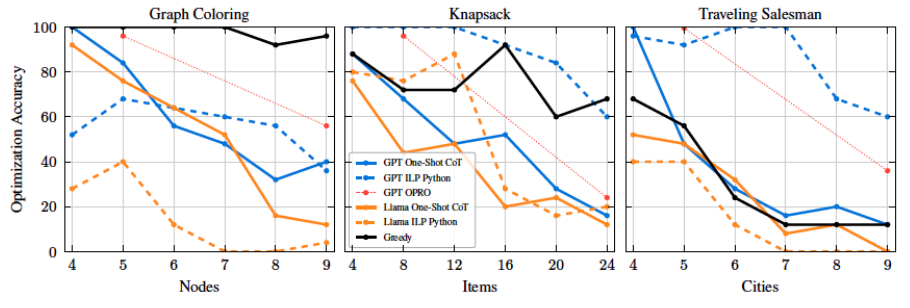}
\caption{LLM and Greedy Heuristic Baseline Optimisation Accuracy on Textbook Variants of Three NP-Hard Tasks, Duchnowski et al. (2025)}
\label{duchnowski-graphs}
\end{figure}

To examine the ability of LLMs to generalise to new versions of the three optimisation tasks, \cite{Duchnowski&etal2025} test them on reformulations in natural 
language, that diverge from the specifications of the problems given in textbooks. These reformulations present the tasks in ways that the models have not encountered 
in their training data. \cite{Duchnowski&etal2025} experiment with one-shot chain of thought  (CoT) prompting for these models, instructing them to reproduce
the inference steps that they have applied in solving the problem. They also compare several of the models to enriched variants that translate each task into an 
Integer Linear Program (ILP) encoding in Python, and apply a Gurobi solver to it.\footnote{These enriched models are an instance of what \cite{Lappin2025b} 
  describes as a \emph{federative} neuro-symbolic architecture, in which the output of an unenriched DNN is passed to a symbolic reasoning component for 
  inference. \cite{Cunnington&etal2023} use this architecture for complex image interpretation.} They use a greedy heuristic algorithm, which selects locally optimal 
choices, as a baseline system. \cite{Duchnowski&etal2025}'s graphs in Figure \ref{duchnowski-graphs} show the accuracy of different versions of ChatGPT-4o
and Llama 3.1, as well as the greedy heuristic algorithm, relative to the number of items in the task, for each of the three problems.

These results indicate that the accuracy of the models declines quickly with the increase of items in the task. Python ILP ChatGPT-4o is the best 
performing system overall, and Python ILP Llama 3.1 does not do particularly well on any of the tasks. Interestingly, the greedy heuristic algorithm does as
well or better than most of the LLM models, for all three tasks, although Python ILP ChatGPT-4o outperforms it (and the other models) in the knapsack and
the travelling salesman tasks. \cite{Duchnowski&etal2025} do not test human subjects on these tasks, and so we cannot compare the performance of their 
models to that of humans. 

If we take experimenting with variation in the form in which an NP-hard optimisation task is formulated as a species of out of domain testing, then 
\cite{Duchnowski&etal2025}'s work supports the view that LLMs have difficulty with generalised real world reasoning. They are efficient at pattern
recognition. They can be fine-tuned or prompted for few shot learning in domain transfer. However, they do not do well on reasoning for
hard optimisation problems across domains, even when enriched with symbolic inference components, or through chain of thought prompting. 

\section{Conclusions and Future Work}

The experimental work considered here strongly indicates that the more extreme views of the linguistic and reasoning abilities of LLMs which are frequently
expressed in current discussions of AI are not well motivated by the facts. LLMs are able to equal or surpass human performance on many cognitively 
interesting linguistic tasks, but the way in which they process and represent linguistic information is quite different. They are by no means simply "returning
their training data". They can recognise complex syntactic structures and semantic relations that extend far beyond those that they have encountered in
training. They often surpass human performance on these tasks. However, they do not exhibit human cohesion or originality in discourse. Careful
examination of  LLM generated narratives for sequences of visual scenes shows that these lack the underlying coherence and creativity that characterise 
human discourse. LLM narrative fluency camouflages a tendency to rely on patterns encountered in extensive training data, to create rich multimodal word 
embeddings. These determine a probability distribution over possible continuations of a discourse describing a situation, given the strings already produced. 
The result is often a verbose, conjunctively descriptive narrative. 

Recent research on inference and planning tasks further supports the view that LLMs and humans process and represent information differently. It also 
undermines the two extreme views of LLMs concerning their reasoning abilities. They are neither stochastic parrots nor agents of general intelligence. 
The models do well on inference and planning problems for which they have been fine-tuned, but their performance declines significantly for out of domain 
testing on these problems. They appear to rely primarily on sophisticated pattern recognition rather than deep, domain general reasoning. 

In order to further clarify the issues discussed in this article it is worth focussing future research on at least three issues. First how can we constrain the training
data and processing mechanisms of LLMs to achieve greater convergence between their linguistic behaviour and that of humans? Success in this line of research 
will illuminate human language acquisition, processing, and representation. Second, modifying LLM design to arrive at greater coherence and surprisal in 
multimodal narrative generation would deepen our insight into what is distinctively human in discourse. Finally, improving few shot training for domain transfer 
in inference and planning will enhance the usefulness of LLMs for tasks that require general reasoning. 

\section{Acknowledgments}

Earlier versions of this paper were presented to the seminar of the Centre for Fundamentals of AI and Computational Theory in the School of Electronic Engineering
and Computer Science at Queen Mary University of London on May 13, 2026, and to the NLP Reading Group of the McGill University Computer Science Department
on May 29, 2026. I am grateful to the audiences of both groups for helpful discussion of the ideas proposed here.

Some of the work reported in this paper was supported by a grant from the Swedish Research Council (VR project 2014-39) for the establishment of the Centre for 
Linguistic Theory and Studies in Probability (CLASP) at the University of Gothenburg. I am grateful to colleagues in CLASP for their cooperation and joint work in doing
some of the research cited here. I bear sole responsibility for the content of this paper, and any errors that it may contain.

No LLM or other AI system was used to generate any of the text of this article.

\bibliographystyle{apalike}
\bibliography{lappin_arxiv}

\end{document}